%% file: main.tex
\documentclass[10pt,twocolumn,letterpaper]{article}

\usepackage{iccv}      


\definecolor{iccvblue}{rgb}{0.21,0.49,0.74}
\usepackage[pagebackref,breaklinks,colorlinks,allcolors=iccvblue]{hyperref}
\usepackage{multirow}
\usepackage{graphicx}
\usepackage{amsmath}
\usepackage{amssymb}
\usepackage{booktabs}
\usepackage{svg}

\def\paperID{2564} 
\def\confName{ICCV}
\def\confYear{2025}

\title{OracleFusion: Assisting the Decipherment of Oracle Bone Script \\with
Structurally Constrained Semantic Typography}

\author{Caoshuo Li\textsuperscript{1}
, Zengmao Ding\textsuperscript{3}
, Xiaobin Hu\textsuperscript{2}
, Bang Li\textsuperscript{3}
, Donghao Luo\textsuperscript{2}
, AndyPianWu\textsuperscript{2}, \\
Chaoyang Wang\textsuperscript{2}, Chengjie Wang\textsuperscript{2}, 
Taisong Jin\textsuperscript{1}
, 
SevenShu\textsuperscript{2}, 
Yunsheng Wu\textsuperscript{2},
Yongge Liu\textsuperscript{3},
Rongrong Ji\textsuperscript{1}
\\
\textsuperscript{1}Xiamen University \textsuperscript{2}Tencent \textsuperscript{3}Anyang Normal University \\
{\tt\small \ licaoshuo@stu.xmu.edu.cn, \{jintaisong, rrji\}@xmu.edu.cn,} {\tt\small \{xiaobinhu, michaelluo\}@tencent.com} \\
{\tt\small \textbf{Project page:\url{https://github.com/lcs0215/OracleFusion}}}\\
}

\begin{document}
\maketitle

\input{sec/0_abstract}
\input{sec/1_intro}
\input{sec/2_related_works}
\input{sec/3_methods}

\input{sec/4_experiments}
\input{sec/5_conclusion}

{
    \small
    \bibliographystyle{ieeenat_fullname}
    \bibliography{main}
}
\input{supplement}

\end{document}

%% file: sec/0_abstract.tex
\begin{abstract}
As one of the earliest ancient languages, Oracle Bone Script (\textbf{OBS}) encapsulates the cultural records and intellectual expressions of ancient civilizations. Despite the discovery of approximately 4,500 OBS characters, only about 1,600 have been deciphered. The remaining undeciphered ones, with their complex structure and abstract imagery, pose significant challenges for interpretation. To address these challenges, this paper proposes a novel two-stage semantic typography framework, named \textbf{OracleFusion}. In the first stage, this approach leverages the Multimodal Large Language Model (MLLM) with enhanced Spatial Awareness Reasoning (SAR) to analyze the glyph structure of the OBS character and perform visual localization of key components. In the second stage, we introduce Oracle Structural Vector Fusion (\textbf{SOVF}),  incorporating glyph structure constraints and glyph maintenance constraints to ensure the accurate generation of semantically enriched vector fonts. This approach preserves the objective integrity of the glyph structure, offering visually enhanced representations that assist experts in deciphering OBS. Extensive qualitative and quantitative experiments demonstrate that OracleFusion outperforms state-of-the-art baseline models in terms of semantics, visual appeal, and glyph maintenance, significantly enhancing both readability and aesthetic quality. Furthermore, OracleFusion provides expert-like insights on unseen oracle characters, making it a valuable tool for advancing the decipherment of OBS.
\end{abstract}

%% file: sec/1_intro.tex
\vspace{-0.5em}
\section{Introduction}
\label{sec:intro}
Oracle bone script (OBS) is one of the earliest ancient languages, originating approximately 3,000 years ago during the Shang Dynasty. It was inscribed on turtle shells and animal bones to record significant events, divinations, and rituals, holding considerable historical and cultural significance. This ancient script provides invaluable insights into early Chinese civilization and represents a foundational stage in the evolution of modern Chinese characters. However, despite discovering approximately 4,500 OBS characters, only around 1,600 have been deciphered, leaving this ancient language shrouded in mystery.
\begin{figure}[t]
    \centering
    \vspace{-1em}
    \includegraphics[width=1.0\linewidth]{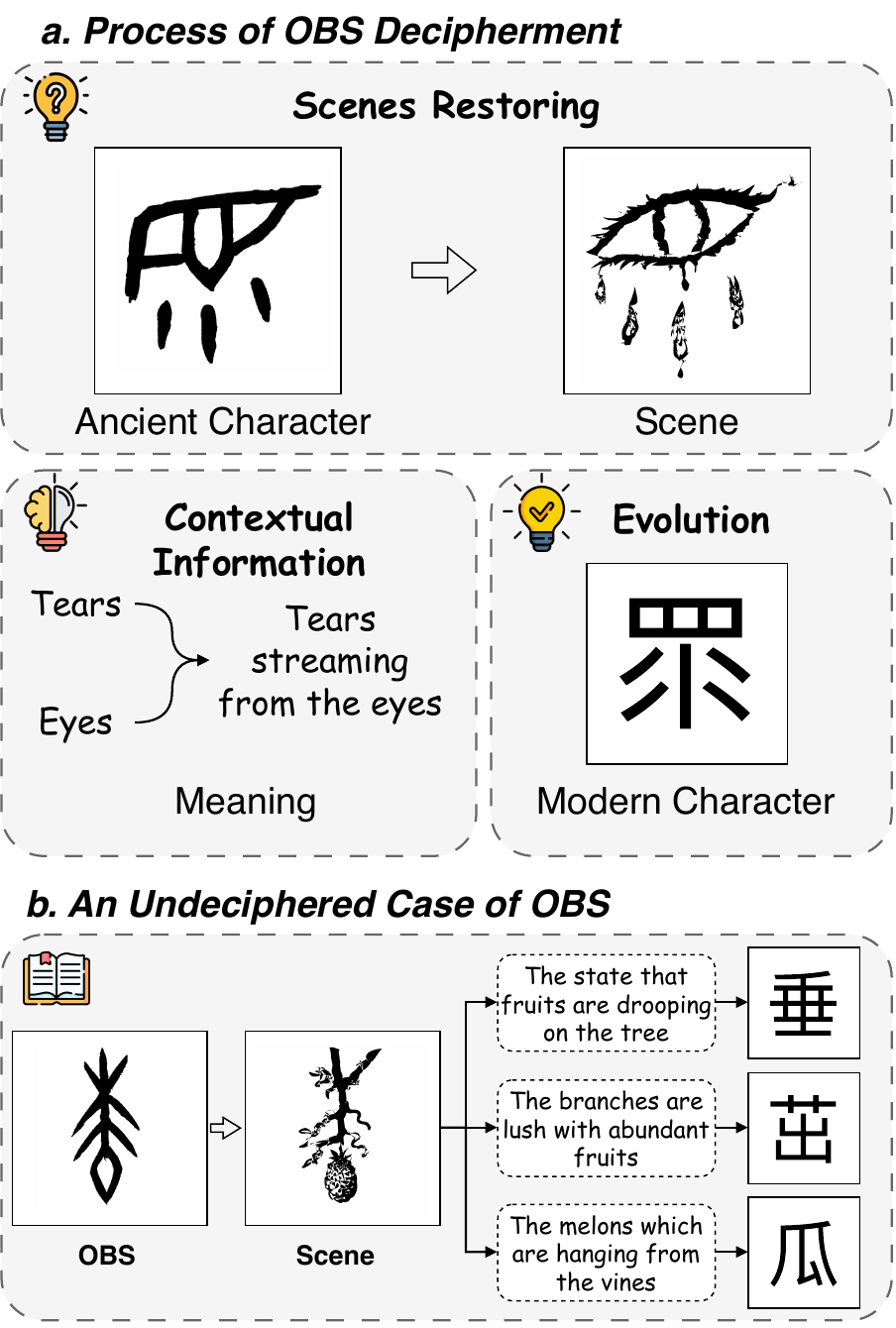}
    \vspace{-1.75em}
    \caption{Illustration of OBS decipherment process. }
    \label{fig:decipherment}
    \vspace{-1.75em}
\end{figure}
The decipherment of OBS is an exceedingly intricate process that requiring a comprehensive consideration of the morphological composition of the characters and their contextual relationships. As depicted in Fig.~\ref{fig:decipherment}(a), this process typically begins with a thorough analysis of the character forms to reconstruct the scenes these ancient symbols represent. Subsequently, by scrutinizing the surrounding context, researchers infer the specific meanings embedded within these inscriptions. Ultimately, through a synthesis of pronunciation and the evolutionary transformations of character forms, the corresponding modern Chinese characters are identified.

Recent research has employed deep learning models to tackle the complex task of deciphering OBS. A notable example is the OBS Decipher (OBSD)~\cite{obsd}, which utilizes conditional diffusion models to generate modern Chinese representations of ancient OBS characters. 
However, OBSD has its limitations. As an end-to-end system, it lacks explicit semantic analysis of OBS characters, reducing interpretability. As depicted in Figure~\ref{fig:decipherment}(b), an undeciphered OBS character often possesses multiple potential meanings, making decipherment an exploration of possible semantic interpretations of the character’s structure. This absence of semantic analysis limits OBSD’s effectiveness, providing minimal insight into the cultural and historical nuances essential for comprehensive OBS decipherment.

Deciphering OBS poses a significant challenge in applying contemporary AI tools, such as LLMs and diffusion models, to explore a more interpretable approach. A core difficulty lies in systematically analyzing the structural composition of OBS glyphs and elucidating their component meanings, which is essential for generating insights that support expert decipherment. Equally critical is reconstructing the scenes and actions represented by OBS characters while preserving the original glyph structure, a task fundamental to studying the evolution of ancient scripts.

To address the aforementioned challenges, we propose OracleFusion, a novel two-stage semantic typography framework. In the first stage, we decompose the task into two sub-problems: oracle bone semantics understanding and oracle bone visual grounding, leveraging the advanced vision-language reasoning capabilities of MLLM to analyze oracle glyph structures and localize key components. We further introduce Spatial Awareness Reasoning (SAR) to enhance MLLM’s spatial comprehension and structural reasoning. In the second stage, we introduce Structural Oracle Vector Fusion (SOVF). Given the complexity of oracle bone glyphs, SOVF enforces structural constraints to ensure that generated objects align with their intended semantics. Specifically, SOVF utilizes oracle glyph structural information to regulate spatial generation, ensuring radicals are positioned correctly. To further preserve glyph integrity, SOVF incorporates SKST loss, maintaining structural fidelity. OracleFusion demonstrates oracle character comprehension, reconstructing scenes of ancient characters, providing valuable insights to assist experts in deciphering oracle bone scripts. Furthermore, we propose RMOBS (Radical-Focused Multimodal Oracle Bone Script), a curated dataset supporting research on oracle glyph structures and semantic representations.

The main contributions are summarized  as:
\begin{itemize}[topsep=3pt, partopsep=3pt, itemsep=3pt]
    \item We propose \textbf{OracleFusion}, a novel two-stage semantic typography framework for generating semantically rich vectorized fonts that accurately convey the essence of the original glyphs, providing a structured visual reference for oracle character interpretation. Extensive experiments demonstrate the superior reasoning and generation capabilities of OracleFusion, validating its effectiveness in supporting OBS decipherment.
    \item In the first stage, we leverage MLLM's advanced vision-language reasoning to analyze oracle glyph structures, understand oracle semantics, and localize key components. To further enhance spatial comprehension, we introduce Spatial Awareness Reasoning (SAR), enabling MLLM to model the relative positioning of key components, ensuring a more structured and interpretable representation of oracle glyphs.
    
    \item In the second stage, we introduce \textbf{SOVF}, a test-time training, glyph structure-constrained semantic typography method that enforces constraints based on oracle glyph structures, ensuring the generated objects align with both semantic integrity and spatial accuracy while effectively maintaining the original glyph structure.
    \item We propose \textbf{RMOBS}, a high-quality multimodal dataset consisting of over 20K OBS samples across 900 characters. Each sample is enriched with radical-level structural annotations, ensuring precise structured interpretation through data refinement and expert validation.

\end{itemize}

%% file: sec/2_related_works.tex
\vspace{-0.5em}
\section{Related Works}
\label{sec:Related_Works}
\subsection{Oracle Character Processing}
Deep learning is integral to OBS processing. Some studies~\cite{oraclepoints, oracle1, oracle2, oracle3} focus on character recognition, denoising, and image translation. Others, like Genov~\cite{genov}, employ VLMs to expand the semantic context via visual guidance. More recent work targets OBS decipherment. OBS Decipher (OBSD)~\cite{obsd} leverages conditional diffusion models to generate modern Chinese counterparts, aiding interpretation. However, OBSD’s end-to-end design limits interpretability, as it lacks explicit semantic analysis, constraining its ability to capture cultural and historical context essential for comprehensive decipherment.

\subsection{Conditional Vector Graphics Generation}
Recent advancements in image and video generation technology have progressed at a remarkable pace \cite{hu2024diffumatting, ji2025sonic, hu2020face}. 
Large-scale text-to-image diffusion models like Stable Diffusion~\cite{sd}, GLIDE~\cite{glide}, Imagen~\cite{imagen}, and DALL\(\cdot\)E 2~\cite{dalle2} have powerful prior capabilities that drive research in text-guided vector image generation. Approaches like VectorFusion~\cite{vectorfusion} and Diffsketcher~\cite{diffsketcher} successfully applied Score Distillation Sampling (SDS) loss~\cite{sds} to vector generation tasks. In semantic typography, Word-As-Image~\cite{wordasimage} combines the semantic proximity enforced by SDS loss with glyph preservation. DS-Fusion~\cite{DS_Fusion} leverages diffusion models as discriminators, using adversarial learning for semantic font stylization. However, due to the complexity of oracle glyphs, such methods struggle to handle radicals. Conditional controllability for raster images has been explored by layout-to-image models like GlIGEN~\cite{gligen}, LDM~\cite{LDM}, and BoxDiff~\cite{box_diff} (\emph{e.g.}, with bounding boxes and masks). Yet, layout-based vector generation remains largely unexplored.

\begin{figure}[t]
    \centering
    \vspace{-0.5em}
    \includegraphics[width=1.0\linewidth]{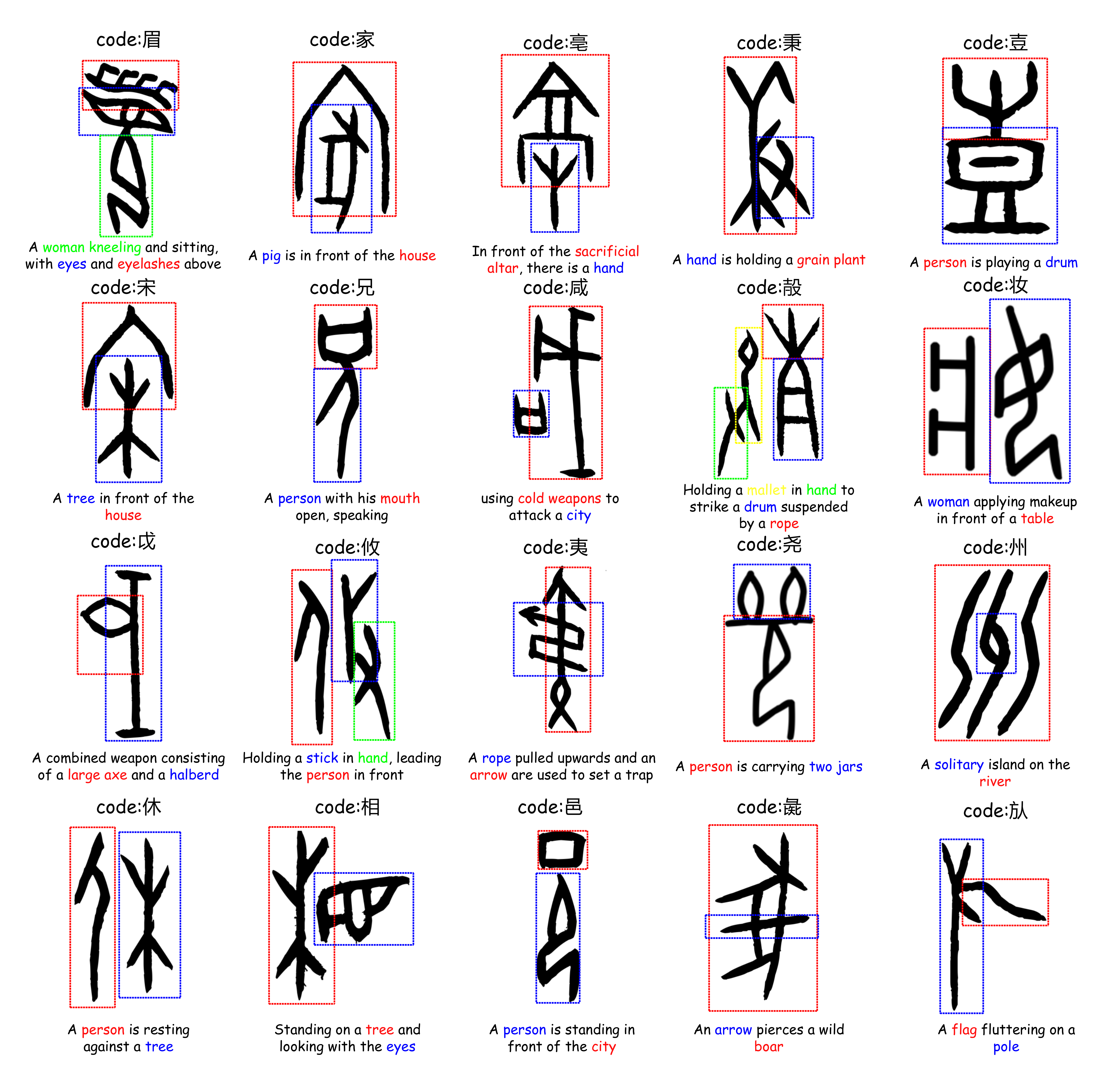}
    \vspace{-2em}
    \caption{Visualization of RMOBS annotations, including oracle glyph \(o_i\), semantic concept \(T_{gi}\), bounding box layout \(L_{gi}\), and key component \(K_{gi}\). Each bounding box is color-matched to its corresponding key component.}
    \label{fig:dataset}
    \vspace{-1.5em}
\end{figure}

\subsection{Multimodal Large Language Models}
Incorporating grounding information into LLMs~\cite{llm_ground} often enhances their reasoning capabilities, and this approach has been applied across many fields \cite{hu2022autogan, hu2023high}. Some studies~\cite{visorgpt, LDM, layoutgpt, controlgpt, shen2025align} have explored using LLMs to plan layouts for image generation. For instance, LayoutGPT~\cite{layoutgpt} have focused on generating structured layouts for infographics and diagrams, where spatial arrangement is critical. However, these methods are primarily designed for natural images or straightforward graphic representations and struggle to adapt to OBS data due to the complex structure, symbolic meanings, and historical context inherent in OBS.


%% file: sec/3_methods.tex
\section{Comparisons with Other Approaches}
\noindent \textbf{Differences against GenOV~\cite{genov}. }1) \emph{Structural Decipherment}: GenOV, based on ControlNet, generates photographic images, while OracleFusion, based on SOVF, produces SVG vector pictograms that better preserve both structural information and original glyphs, facilitating the decipherment of OBS. 2) \emph{Black-White Vector Design}: OracleFusion illustrations focus on modifying the oracle's geometry to convey meaning, intentionally omitting color, texture, and embellishments. This ensures simple, concise, black-and-white designs that clearly represent semantics. Furthermore, the vector-based representation enables smooth rasterization at any scale and supports optional style adaptations, such as color and texture adjustments.

\noindent \textbf{Comparision with Word-As-Image~\cite{wordasimage}. }1) \emph{Semantics Representation}: OracleFusion employs GSDS Loss to enforce region-based constraints on the cross-attention mechanism of radicals, ensuring that the generated results semantically align with the target structure. In contrast, Word-As-Image often overlooks certain oracle radicals or places them incorrectly. 2) \emph{Glyph Maintenance}: Our proposed OGV method extracts skeleton control points during vectorization. SKST Loss then constraints these skeleton points, ensuring better preservation of the original glyph shape. In contrast, Word-As-Image only constrains the outer contour, leading to greater deviations from the original glyph.
\section{Method}
In this section, we begin by presenting the data collection and annotation process for our proposed OBS dataset. Subsequently, we provide a comprehensive description of the modules within the OracleFusion framework. The overview of the OracleFusion framework is depicted in Figure~\ref{fig:framework} .
\subsection{Data Collection and Annotations}
\label{sec:dataset}

In this study, we curated RMOBS, a multimodal dataset comprising over 20K samples across 900 deciphered OBS characters with pictographic components. In contrast, GenOV~\cite{genov} only contains 364 characters. To obtain detailed pictographic explanations, we first collected OBS data from the website~\cite{vividict}, which provided comprehensive descriptions of each character’s form and components. Based on these descriptions, we manually annotated bounding boxes and semantic concepts for each component (radicals) in the OBS images, during the data cleaning process, redundant parts were  removed, and key terms were extracted to ensure consistency, accuracy, and clarity in the descriptions. This dataset contains over 20K entries, as illustrated in Figure~\ref{fig:dataset}, with each entry comprising an oracle glyph \(o_i\), a concept \(T_{gi}\), the key components \(K_{gi}\) and a bounding box layout \(L_{gi}\). Formally, this dataset is represented as \(\mathcal{O} = \{(o_i, T_{gi}, K_{gi}, L_{gi})\}_{i=1}^{20K}\).

\subsection{Preliminaries: Score Distillation Sampling}

DreamFusion~\cite{dreamfusion} utilizes Score Distillation Sampling (SDS) to optimize NeRF parameters for text-to-3D generation, leveraging pretrained diffusion models. The SDS approach introduces a differentiable function 
\(R\) that renders an image \(x=R(\theta)\)
, where \(\theta\) represents the NeRF parameters. During each iteration, \(x\) is perturbed by noise at a specified diffusion step, generating a noisy image 
\(z_t = \alpha_tx + \sigma_t\epsilon\). This noised image is processed by a pretrained U-Net~\cite{unet} to predict the noise 
\(\epsilon\), allowing the model to backpropagate gradients to refine the NeRF parameters. The SDS loss gradient for a pretrained model \(\epsilon_{\phi}\) is defined as:
\vspace{-0.5em}
\begin{equation}
\nabla_\theta \mathcal{L}_{\text{SDS}} = w(t) \left( \epsilon_\phi(z_t(x); y, t) - \epsilon \right) \frac{\partial x}{\partial \theta} ,
\end{equation}
where \(y\) is the text prompt, 
\(w(t)\) is a weighting function based on 
\(t\), and \(z_t(x)\) represents the noised image. 

VectorFusion~\cite{vectorfusion} and Word-As-Image~\cite{wordasimage} both utilize SDS loss for text-to-SVG generation. We followed the technical steps in ~\cite{vectorfusion, wordasimage}, \textit{i.e.}, we add a suffix to the prompt and apply augmentation to images generated by DiffVG~\cite{diffvg}. The SDS loss is subsequently applied in the latent space of UNet in Stable Diffusion Model~\cite{sd} to update the SVG parameters. The SDS loss is defined as:
\vspace{-0.5em}
\begin{align}
\nabla_\theta & \, \mathcal{L}_{\text{LSDS}} =  \notag \\ & 
\mathbb{E}_{t, \epsilon} \Bigg[   w(t) \Big( \hat{\epsilon}_\phi \left(\alpha_t z_t + \sigma_t \epsilon, y\right) - \epsilon \Big)
\frac{\partial z}{\partial x_{\text{aug}}} \frac{\partial x_{\text{aug}}}{\partial \theta} \Bigg].
\label{lsds}
\end{align}
\subsection{Oracle Glyph Vectorization}
\label{sec:vector}
\begin{figure}[t]
    \centering
    \vspace{-0.5em}
    \includegraphics[width=1.0\linewidth]{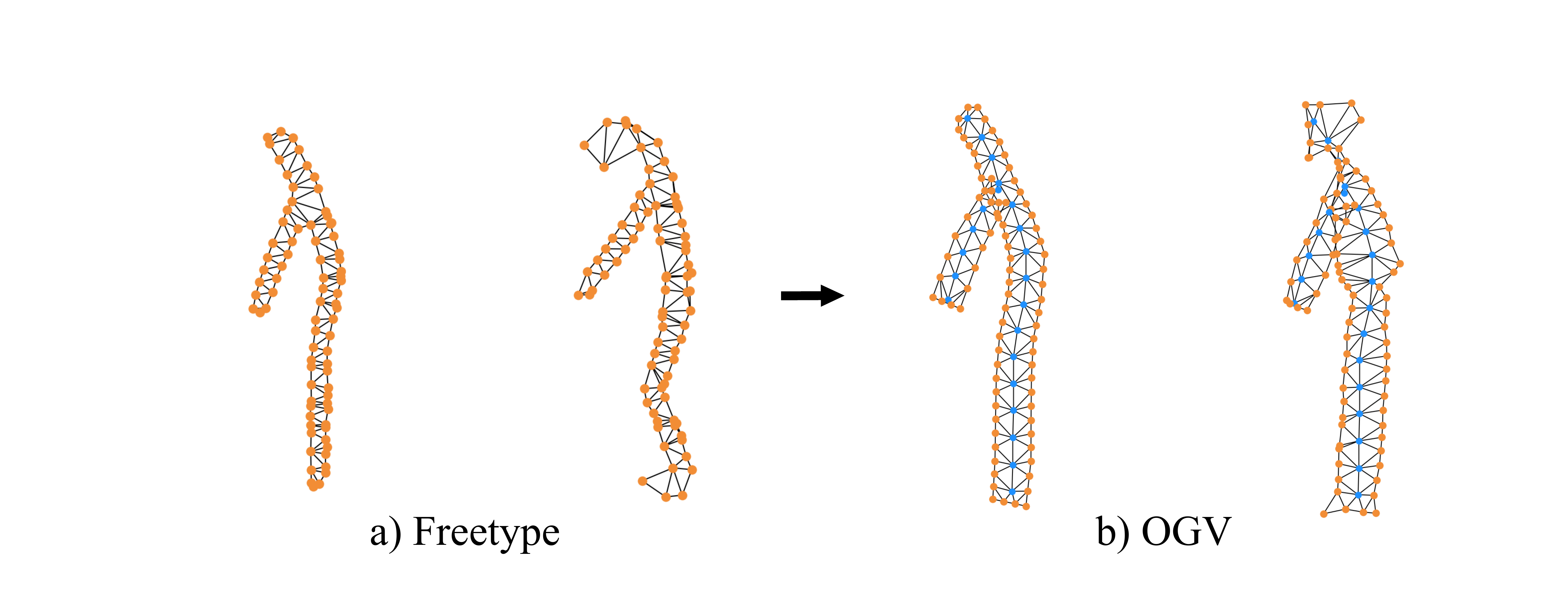
    }
    \vspace{-1.75em}
    \caption{Delaunay triangulation of the initial and resulting points for Freetype and OGV. The orange points are derived from the outline, while the blue points in OGV are derived from the skeleton.}
    \label{fig:skelon}
    \vspace{-1.5em}
\end{figure}

Given the diverse forms of OBS and the limitation of the vector font library—covering only 4K of an estimated 50K OBS images—we propose the Oracle Glyph Vectorization (OGV) method to efficiently process OBS images and extract key stroke features. As shown in Figure~\ref{fig:skelon}, the leftmost image illustrates the Freetype~\cite{freetype} font outline, while the third image presents the outline enhanced by our method with additional control points derived from the skeleton structure. To refine strokes into \(n\) single-pixel-wide skeleton paths, we apply the Zhang-Suen (ZS) skeletonization algorithm~\cite{zs_skeleton}, yielding \( P^s = \{p^s_i\}_{i=1}^n \). Each path \( p^s_i = \{(x_j, y_j)\}^m_j \) consists of points \( C^s_j = (x_j, y_j) \). Direction vectors are computed as \( \mathbf{v}_j = C^s_{j+1} - C^s_j \) and rotated by 90° to obtain normal vectors \( \mathbf{\mu}_j = (-v_{j,y}, v_{j,x}) \). To maintain contour continuity, each normal vector is smoothed using a sliding window of size \( k \), yielding:
\(
\mathbf{\tilde{\mu}}_j = \frac{1}{k} \sum_{i=j-\frac{k}{2}}^{j+\frac{k}{2}} \mathbf{\mu}_i.
\)
The smoothed normals reduce noise and improve contour stability. Given a stroke width \( w \), left and right contour points are defined as \( p^l = \{C^s_j + w\mathbf{\tilde{\mu}}_j\} \) and \( p^r = \{C^s_j - w\mathbf{\tilde{\mu}}_j\} \). Cubic splines are then fitted to \( p^l \) and \( p^r \), generating interpolated points \( \tilde{p}^l \) and \( \tilde{p}^r \). The final contour \( P = \{\tilde{p}^l, \tilde{p}^r\} \) is constructed around each stroke segment. By repeating this process for all \( n \) skeleton paths, the full set of outer contour points \( P = \{p_i\}_{i=1}^n \) is obtained.

\subsection{Deciphering and Grounding of MLLM}
\begin{figure*}
    \centering
    \vspace{-1em}
    \includegraphics[width=1.0\linewidth]{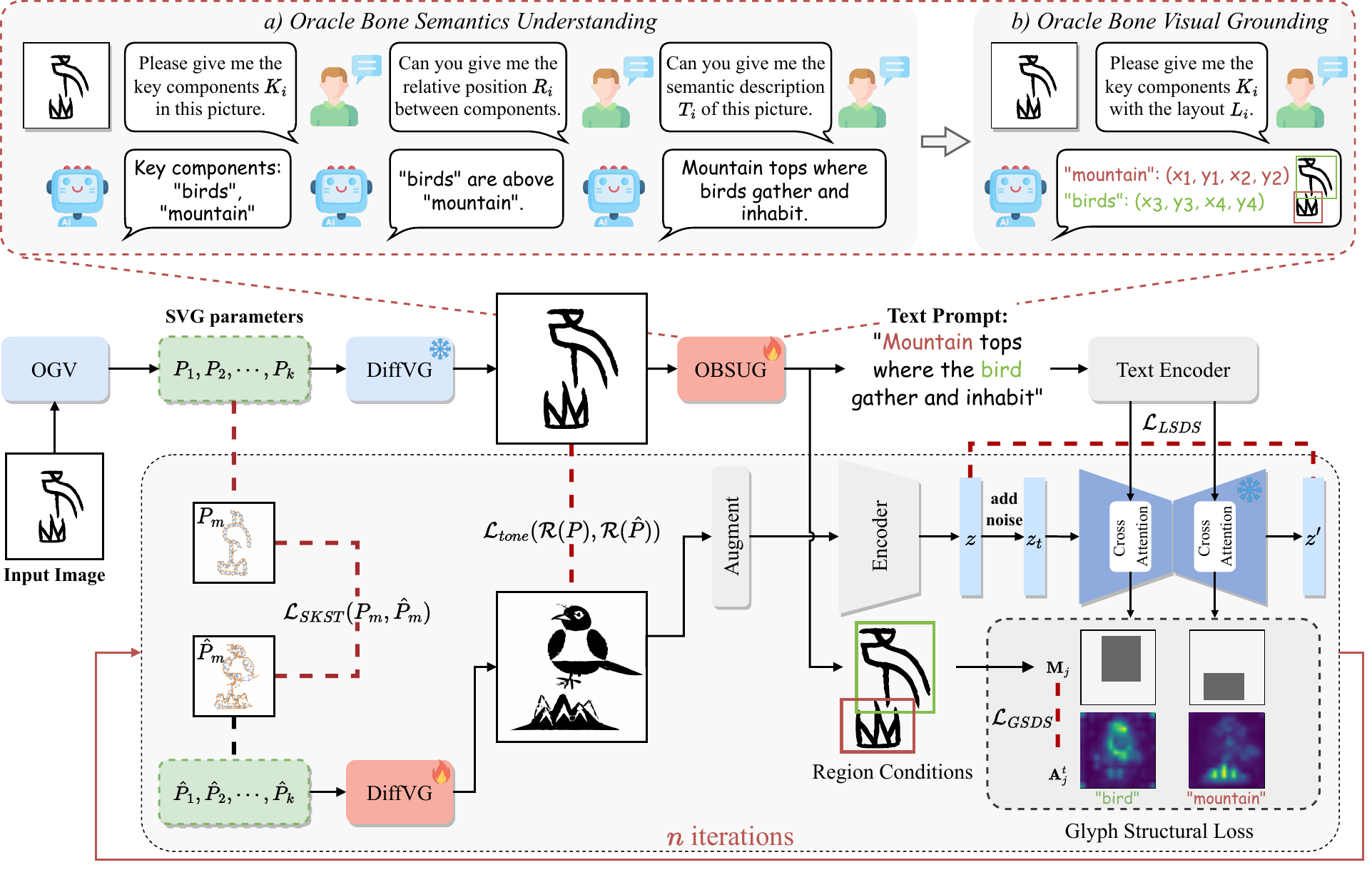}
    \vspace{-2em}
    \caption{Overview of the proposed OracleFusion framework. This framework first employs OGV to transform oracle images into vectorized font and then utilizes OBSUG to generate prompt (oracle semantics) and region conditions, enables vectorized glyph morphing with \(\mathcal{L}_{\text{LSDS}}\) and \(\mathcal{L}_{\text{GSDS}}\) by updating the SVG parameters. \(\mathcal{D}\) represents Delaunay triangulation and LPF represents a low pass filter. OBSUG framework: (a) OBSU: The MLLM identifies key components \(K_i\), outputs relative positional relationships \(R_i\), and generates a description \(T_i\) of the oracle glyph \(o_i\). (b) OBVG: the MLLM grounds each semantic component spatially, assigning layout \(L_i\) to key components \(K_i\).} 
    \label{fig:framework}
    \vspace{-0.5em}
\end{figure*}

To enhance the understanding and utilization of OBS data, we propose the Oracle Bone Semantics Understanding and Visual Grounding (OBSUG) framework. Leveraging the vision-language reasoning capabilities of MLLM~\cite{qwen_vl}, OBSUG aligns abstract oracle bone images with rich semantic information, providing critical insights for decipherment. As shown in Figure~\ref{fig:framework}, OBSUG employs a multi-turn dialogue mechanism to process OBS data.

\noindent{\textbf{Oracle Bone Semantics Understanding (OBSU). }}The OBSU module comprises three sequential sub-stages: key structural components identification, Spatial Awareness Reasoning (SAR) and fine-grained semantics generation. 

In the first sub-stage, the MLLM \(\Psi\) receives the oracle bone glyph \(o_i\) and a task-specific description \(\theta_{key}\), which instructs it to identify key structural components. This process can be defined as:
\(K_i = \Psi(o_i;\theta_{key})
\),
where \(K_i=\{k_1, k_2, \cdots, k_M\}\) represents essential elements of the oracle glyph (\emph{e.g.}, radicals or distinct parts).

In the second sub-stages, we introduce \emph{Spatial Awareness Reasoning} (SAR) to guides model to output the relative positional relationships among the identified key components in the form of a Directed Acyclic Graph (DAG), which can be denoted as:
\vspace{-0.5em}
\begin{equation}
\!\!\!G_i = (V, E),  V = K_i, 
     E = \{(k_a, k_b, r) \mid k_a, k_b \in V\},
\end{equation}
where \( r \) denotes the relative spatial relation between \(K_i\).

The MLLM \( \Psi \) then generates the relative spatial position relationships \(R_i\), which can be denoted as:
\vspace{-0.5em}
\begin{equation}
    R_i = \Psi(o_i,K_i; \theta_{\text{spa}}),
\end{equation}
where \(\theta_{\text{spa}}\) represents the task-specific spatial description.

In the third sub-stage, the model is further prompted with \(K_i\) and a general task description \(\theta_{cap}\) that instructs it to generate an fine-grained semantic interpretation \(T_i\) capturing the overall meaning of \(o_i\). The abstract process can be described mathematically as follows:
\vspace{-0.5em}
\begin{equation}
    T_i = \Psi(o_i, K_i, R_i ; \theta_{cap}).
\end{equation}
\noindent{\textbf{Oracle Bone Visual Grounding (OBVG). }}In the final stage, the model \(\Psi\) performs visual grounding to establish the spatial layout of the identified components within the glyph. The formulation is defined as:
\vspace{-0.5em}
\begin{equation}
    L_i = \Psi(o_i, K_i, R_i, T_i;\theta_{loc}),
\end{equation}
where \(L_i\) is the spatial layout generated in visual grounding stage, indicating the positions of the identified components within the glyph \(o_i\), and \(\theta_{loc}\) represent the task-specific layout description, including requirements and examples.

\subsection{Structural Oracle Vector Fusion}
In this subsection, we propose a training-free structure-constrained semantic typography framework, named Structural Oracle Vector Fusion (SOVF). This approach leverages structural information in oracle glyph to control the spatial position of the  generated objects, ensuring alignment with the glyph's semantics and preserving the initial glyph.

\noindent{\textbf{Cross-Modal Attention.}}
In Stable Diffusion model~\cite{sd}, conditioning mechanisms enable the formation of cross-attention between text tokens and the latent features. Given text tokens \(\tau_\theta(y)\) and latent features \(z_t\), the cross-attention matrix \(\mathbf{A}\) is defined as follows:
\begin{equation}
\!\!\!\!\mathbf{A} = \text{Softmax}(\frac{\mathbf{Q}\mathbf{K}^{\mathrm{T}}}{\sqrt{d}}), \mathbf{Q} = \mathbf{W}_Qz_t,\enspace\mathbf{K} = \mathbf{W}_K\tau_\theta(y),
\end{equation}
where \(\mathbf{Q}\) and \(\mathbf{K}\) represent the projections of text tokens \(\tau_\theta(y)\) and latent features \(z_t\) through learnable matrices \(\mathbf{W}_Q\) and \(\mathbf{W}_K\), respectively. At each timestep \(t\), \(\tau_\theta(y)\) maps to \(N\) text tokens \(\{b_1, \dots, b_N\}\), producing \(N\) spatial attention maps \(\{\mathbf{A}^t_1, \dots, \mathbf{A}^t_N\}\). Following~\cite{attend_and_excite}, we exclude cross-attention between the start-of-text token ([sot]) and latent features before applying Softmax\((\cdot)\). A Gaussian filter further smooths the cross-attention maps spatially. We impose constraints on the cross-attention maps at a resolution of 16\( \times \)16. Given a text prompt with \(N\) tokens \(B = \{b_j\}\) and a layout \(L = \{l_j\}\) for \(M\) components of the target oracle glyph, we obtain the corresponding spatial cross-attention maps \(\mathcal{A}^t = \{\mathbf{A}^t_j\}\). Each layout location \(l_j\) is defined by its top-left and bottom-right coordinates \(\{(x_j^{tl}, y_j^{tl}), (x_j^{br}, y_j^{br})\}\). The layout \(L\) is then converted into binary spatial masks \(\mathcal{M} = \{\mathbf{M}_j\} \in \mathbb{R}^{16 \times 16 \times M}\). To control the synthesis of specific components \(K\) in the oracle glyph \(o_i\), we introduce the \emph{Glyph Structural Constraint}, which applies targeted constraints on \(\mathcal{A}^t\). This constraint guides the gradient of latent features \(z_t\) and propagates through the SVG parameters via backpropagation.

\noindent{\textbf{Glyph Structural Constraint. }}To ensure synthesized objects adhere to the specified layout, we enforce high cross-attention responses within mask regions while suppressing them outside to prevent drift. Accordingly, we define the Inside-Region and Outside-Region constraints as follows:
\vspace{-1em}
\begin{equation}
    \mathcal{L}^{IR}_{b_j} = 1 - \frac{1}{P}\sum \textbf{Topk}(\mathbf{A}^t_j \cdot \mathbf{M}_j,P),
    \label{con:innerbox}
\end{equation}
\begin{equation}
    \mathcal{L}^{OR}_{b_j} = \frac{1}{P}\sum \textbf{Topk}(\mathbf{A}^t_j \cdot (1-\mathbf{M}_j),P),
    \label{con:outerbox}
\end{equation}
\begin{equation}
    \mathcal{L}_{IR} = \sum_{j\in K}\mathcal{L}^{IR}_{b_j},\enspace\mathcal{L}_{OR} = \sum_{j\in K}\mathcal{L}^{OR}_{b_j},
\end{equation}
\begin{figure}[t]
    \centering
    \vspace{-0.5em}
    \includegraphics[width=1.0\linewidth]{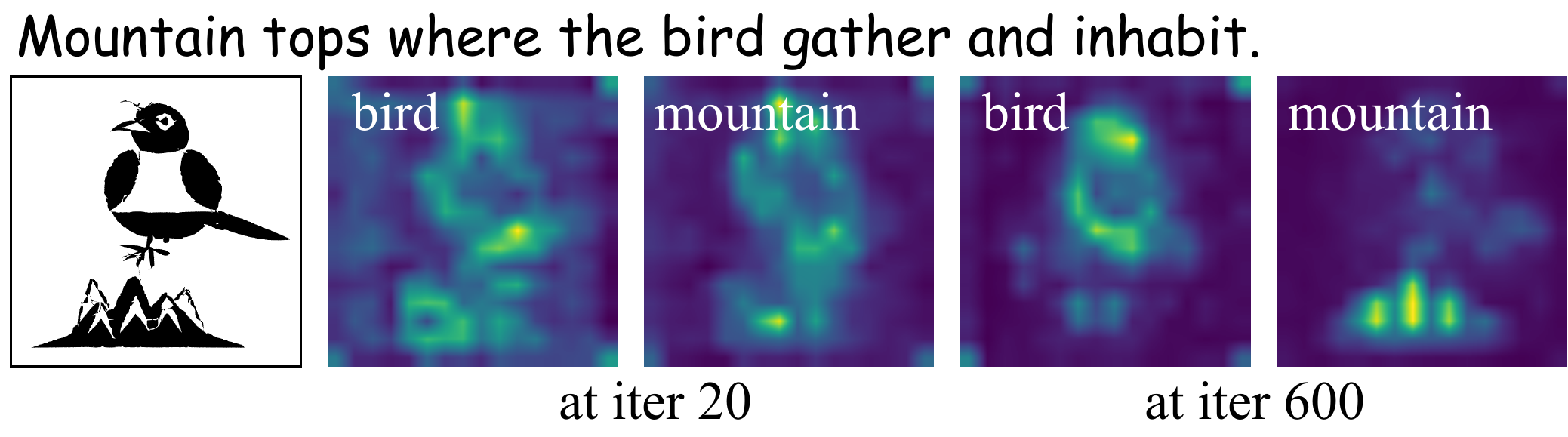}
    \vspace{-2em}
    \caption{Cross-attentions between target tokens \emph{e.g.}, bird, mountain) and latent features of the UNet in Stable Diffusion Model.}
    \label{fig:attention}
    \vspace{-1.5em}
\end{figure}where \(\textbf{Topk}(\cdot,P)\) selects the \(P\) highest cross-attention responses, and \(K\) denotes the set of oracle text tokens for \(M\) essential elements. Our experiments show that constraining only a few high-response elements suffices to influence object synthesis while mitigating constraint impact and preventing denoising failures. Thus, we constrain the \(P\) highest responses to shape gradients and backpropagate onto SVG parameters. The binary mask \(\mathbf{M}_i\) in Eq.~\ref{con:innerbox} maximizes responses within the mask regions, while its complement, \(1-\mathbf{M}_i\), in Eq.~\ref{con:outerbox}, minimizes responses beyond these regions. This complementary mechanism, defined by \(\mathcal{L}_{IR}\) and \(\mathcal{L}_{OR}\), ensures precise object placement in synthesized oracle glyphs. As shown in Figure~\ref{fig:attention}, high-response attention areas, such as the bird and mountain, align perceptually with objects or contextual elements in the synthesized vector fonts. At each timestep, the overall Glyph Structural (GS) Constrained loss is defined as: \(\mathcal{L}_{GS} =  \mathcal{L}_{IR} + \mathcal{L}_{OR}.\)

Similar to the SDS loss~\cite{dreamfusion}, we propose the GSDS loss to update SVG parameters, this optimization process can be formulated as:
\vspace{-1.25em}
\begin{equation}
\nabla_P \mathcal{L}_{\text{GSDS}} = 
\mathbb{E}_{t, \epsilon} \Bigg[   \mathcal{L}_{GS} \frac{\partial z_t}{\partial z} \frac{\partial z}{\partial x_{\text{aug}}} \frac{\partial x_{\text{aug}}}{\partial P} \Bigg].
\end{equation}
\noindent \textbf{Glyph Maintenance. }Our goal is to ensure the synthesized shape conveys the given semantic text using the \(\mathcal{L}_{\text{LSDS}}\) loss in Eq.~\ref{lsds}. However, relying solely on \(\mathcal{L}_{\text{LSDS}}\) can cause significant deviations from the original glyph structure. To address this, we introduce tone loss from~\cite{wordasimage} and introduce the Skeleton Structure (SKST) Loss. Specifically, we extract the skeleton paths \(P^s\) and contour paths \(P\) using OGV in \ref{sec:vector}, denoted as \(P_m = \{P, P^s\}\). Through Delaunay triangulation \( \mathcal{D} \)~\cite{delaunay1934sphere}, we construct a set of triangles \( \mathcal{S} = \{s_{i, j}\}_{i=1}^{n_j} \), where each skeleton point \(C^s_j = p^s_{i,j}\) is linked to \(n_j\) triangles. Within each triangle, the vector from a skeleton point \(C^s_j\) to its connected contour point \(C_i\) is defined as \( \vec{\alpha_{i,j,k}} = C_{i,k} - C^s_j \), where \(j\) indexes the skeleton point, \(i\) the control point, and \(k\) the vertices linking them. The SKST Loss enforces angular consistency between the generated shape \(\hat{P}\) and the original \(P\), minimizing deviations between adjacent points and aligning \( \vec{\alpha_{i,j,k}} \) with its counterpart in the original glyph. This constraint preserves local and global structure, ensuring the synthesized shape remains faithful to the original. The final formulation is:
 \vspace{-0.7em}
\begin{equation}
    \mathcal{L}_{SKST}(P_m, \hat{P}_m) = \frac{1}{N}\sum_{i, j, k} \text{ReLU} \left(- \cos \theta_{i,j,k} \right),
\end{equation}
\vspace{-1em}
\begin{equation}
    \cos \theta_{i,j,k} = \frac{\vec{\alpha_{i,j,k}} \cdot \hat{\vec{\alpha_{i,j,k}}}}{\|\vec{\alpha_{i,j,k}}\| \|\hat{\vec{\alpha_{i,j,k}}}\|},
\end{equation}
where \( N \) represents the total number of vectors in the set. This loss function thus preserves the structural fidelity of the generated glyph by enforcing angle preservation and directional alignment with the original skeleton structure.

Our final objective is then defined by the weighted average of the four terms:
\vspace{-0.5em}
\begin{equation}
    \mathop{\text{min}}\limits_P \nabla_P \mathcal{L}_{\text{LSDS}} + w \cdot \nabla_P\mathcal{L}_{\text{GSDS}} + \beta \cdot \mathcal{L}_{SKST} + \gamma_t \cdot \mathcal{L}_{tone},
\end{equation}
where \(w\) is a learnable weight parameter, \(\beta = 0.5\) and \(\gamma_t\) depends on the step \(t\) as described next. 
\begin{figure*}
    \centering
    \vspace{-0.5em}
    \includegraphics[width=1.0\linewidth]{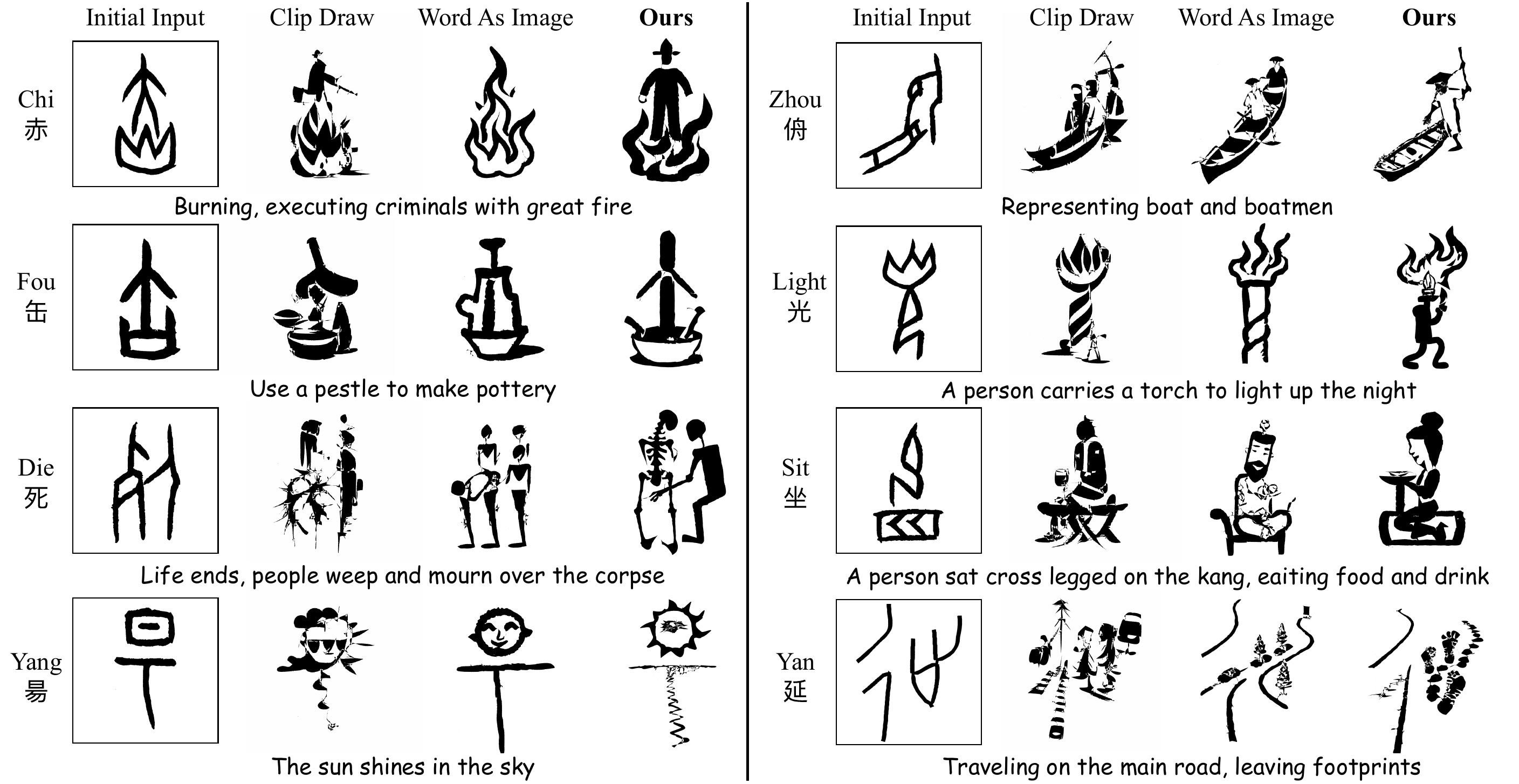}
    \vspace{-2.25em}
    \caption{Qualitative comparison of our OracleFusion with other popular methods. On the left are the initial glyph as input. The results from left to right are obtained by Word-As-Image~\cite{wordasimage}, ClipDraw~\cite{clipdraw} and our OracleFusion, respectively. }
    \label{fig:comparison}
    \vspace{-0.5em}
\end{figure*}

%% file: sec/4_experiments.tex
\section{Experiments}

\subsection{Experiment Settings}
\noindent \textbf{Dataset. }Our experiments are conducted on RMOBS dataset introduced in 
Sec.~\ref{sec:dataset}. We partition the dataset into train and test sets with a ratio of \(9:1\).

\noindent \textbf{Implementation. }The training configuration for OBSUG involved fine-tuning QWEN-VL~\cite{qwen_vl} using LoRA~\cite{lora} with Supervised Fine-Tuning (SFT) at a learning rate of \(1e^{-4}\). The OBSUG was trained on 8 NVIDIA Tesla V100 GPUs, with data handling and acceleration managed by the Swift~\cite{swift} tool. For OBS characters lacking vector fonts, we uniformly apply the vectorization algorithm proposed in Sec.~\ref{sec:vector}. 
For SOVF, we repeat the optimization process for 800 iterations on one RTX3090 GPU.


\begin{figure}[t]
    \centering
    \vspace{-1.25em}
    \includegraphics[width=1.0\linewidth]{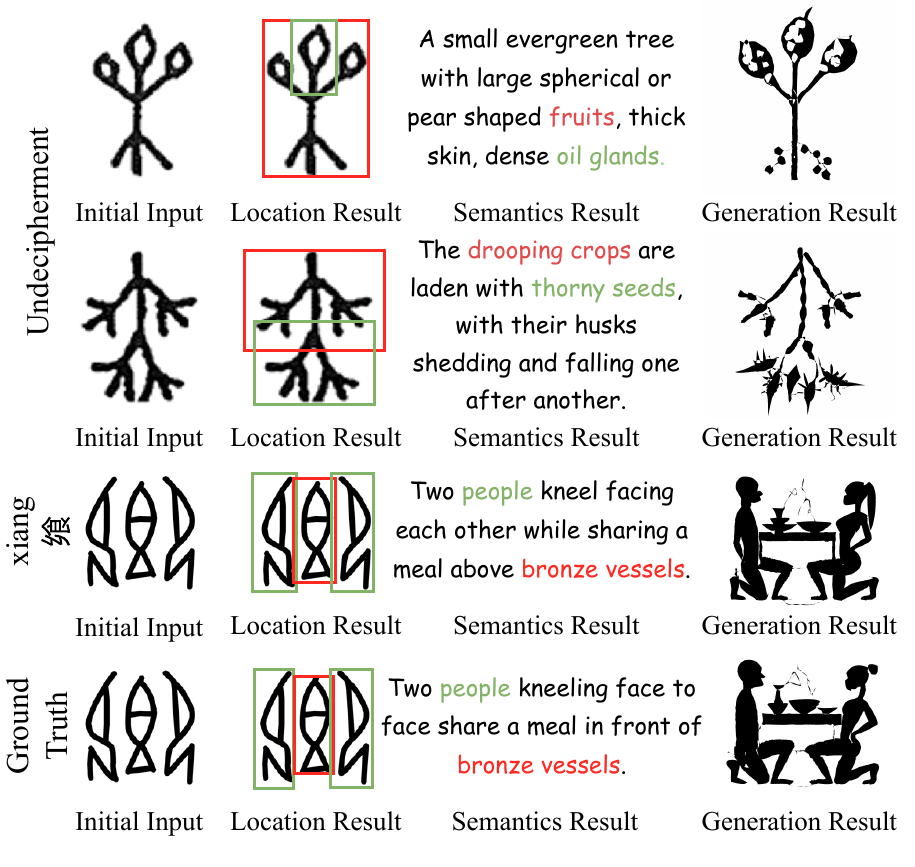}
    \vspace{-2em}
    \caption{Results of unseen undeciphered and deciphered OBS.}
    \label{fig:poyi}
    \vspace{-1.8em}
\end{figure}
\noindent \textbf{Evaluation metrics.} For the ablation study of OBSUG, we utilize BLEU-4~\cite{bleu} and mIoU~\cite{pascl_voc} to assess the semantics representation and grounding ability, respectively. 
To evaluate the semantic typography capabilities of OracleFusion, we assess semantic relevance using CLIPScore~\cite{clipscore}. For glyph maintenance, we designed a glyph matching algorithm, termed Distance, to measure similarity between oracle glyphs. In addition, we conduct a user study to assess the semantics, visual appeal and glyph maintenance. 
\subsection{Decipherment Results}
To evaluate the decipherment capabilities of OracleFusion, we applied it to both undeciphered and deciphered OBS, as shown in Figure~\ref{fig:poyi}. In the first two rows, OracleFusion generates location, semantic, and visual results for an undeciphered character, providing plausible interpretations, such as “drooping crops” and “thorny seeds,” that could offer valuable clues for expert analysis. In the last two rows, OracleFusion accurately replicates the structure and meaning of a deciphered character, closely matching the ground truth. These results demonstrate its effectiveness in capturing both the semantic essence and structural details of OBS, supporting its potential as a valuable tool in OBS decipherment.

\begin{table}[t]
\vspace{-1.15em}
\centering
\small
\renewcommand{\arraystretch}{1.0}
\setlength{\tabcolsep}{1.4pt}{
\begin{tabular}{c c c c c c}
\toprule[0.8pt]
\multirow{2}{*}{Methods} & \multirow{2}{*}{CLIPScore $\uparrow$} & \multirow{2}{*}{Distance $\downarrow$} & \multicolumn{3}{c}{User Study}\\ \cline{4-6} 
  &  & & SR $\uparrow$ & VA$\uparrow$ & GM $\uparrow$\\
\midrule
ClipDraw~\cite{clipdraw} & 27.78 & 1.05  & 3.15  & 3.19 & 3.17  \\
Word-As-Image~\cite{wordasimage} & 27.28 & 0.92  &  3.42  & 3.48 & 3.44 \\
\textbf{OracleFusion (ours)} & \textbf{28.30} & \textbf{0.86}  &  \textbf{3.97}  & \textbf{3.90} & \textbf{3.95} \\
\bottomrule[0.8pt]
\end{tabular}
\vspace{-1em}
}
\caption{Quantitative comparison bewteen our and other methods on CLIPScore, Distance, and user study for Semantics Representation (SR), Visual Appeal (VA) and Glyph Maintenance (GM). }
\label{table:comparison2}
\end{table}
\begin{table}[t]
\vspace{-1em}
\centering
\small
\renewcommand{\arraystretch}{1.0}
\setlength{\tabcolsep}{5.5pt}{
\begin{tabular}{c c c c c}
\toprule[0.8pt]
\multirow{2}{*}{Methods} & \multirow{2}{*}{Acc (\%) $\uparrow$} &  \multirow{2}{*}{mIoU (\%) $\uparrow$} &\multicolumn{2}{c}{BLEU-4~\cite{bleu} $\uparrow$} \\ \cline{4-5} 
  & & & Radical & Holistic\\
\midrule
End-to-End & 81.03  & 72.56 & 0.843 & 0.876\\
Multi-Turn & 81.65 & 72.03 & 0.865 & 0.910 \\
+SAR & \textbf{82.02}  & \textbf{73.94} & \textbf{0.876} & \textbf{0.921} \\
\bottomrule[0.8pt]
\end{tabular}
\vspace{-1em}
}
\caption{Ablation study of OBSUG. Acc denotes the accuracy of the predicted key components count.}
\label{table:OBSUG_Ablation}
\vspace{-2em}
\end{table}
\subsection{Qualitative Results}
The qualitative comparison results are shown in Figure~\ref{fig:comparison}, from which several conclusions can be drawn. While Clipdraw conveys the intended meanings of OBS, its results exhibit sharp, unsmooth edges that deviate significantly from the original glyph structures. In contrast, Word-As-Image better preserves the initial glyph shapes while conveying semantic concepts. However, for complex meanings and structural components, Word-As-Image struggles to represent radicals linked to specific semantic elements. This limitation is evident in “Light” and “Yan.” In “light” glyph, representing a person carries a torch, Word-As-Image omits the person concept. Similarly, in “Yan” glyph, which includes a road and footprints, it overlooks the footprints. In contrast, OracleFusion accurately represents both elements, preserving the original glyph contours while reflecting distinct semantic concepts, proving valuable for studying OBS.

\begin{figure}[t]
    \centering
    \vspace{-1.3em}
    \includegraphics[width=1.0\linewidth]{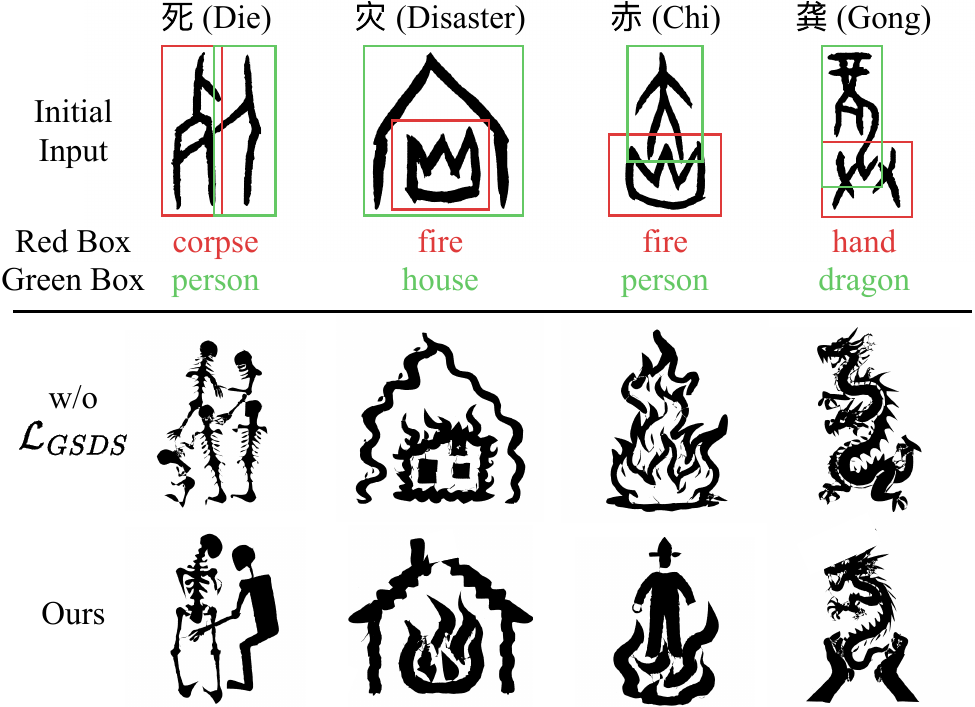}
    \vspace{-2em}
    \caption{Ablation study on w/wo GSDS loss \(\mathcal{L}_{\text{GSDS}}\).}
    \label{fig:ablation_box}
    \vspace{-1.5em}
\end{figure}\subsection{Quantitative Results}
As shown in Table~\ref{table:comparison2}, our method achieves the highest scores in CLIPScore~\cite{clipscore}, indicating that OracleFusion surpasses both CLipDraw and Word-As-Image conveying semantic representation. In terms of glyph maintenance, our method records the lowest distance to the original glyph, highlighting OracleFusion's ability to preserve glyph structure. Additionally, to assess human perception of our generated results, we conduct a user study with 70 participants who are familiar with the principles of Chinese character formation.  This study compares our method with Word-As-Image and ClipDraw on 28 randomly selected OBS, where participants are asked to rate each method from 1 to 5 on Semantics, Visual Appeal, and Glyph Maintenance. Results of the user study show that OracleFusion consistently outperforms other methods across all three dimensions, demonstrating the superiority of our method.

\subsection{Ablation Study}
\noindent\textbf{OBSUG. }
As shown in Figure~\ref{table:OBSUG_Ablation}, the multi-turn approach improves radical count accuracy and BLEU-4~\cite{bleu} scores over end-to-end learning but slightly reduces mIoU. SAR further enhances all metrics, with the most notable mIoU gain, underscoring its role in refining spatial consistency and visual grounding for radicals. These results indicate that iterative reasoning benefits textual alignment, while spatial awareness is crucial for precise structural modeling.

\noindent\textbf{Glyph Structural Constrained loss. }As shown in Figure~\ref{fig:ablation_box}, the results without \(\mathcal{L}_{\text{GSDS}}\) fails to convey the complete concept in all four cases. For instance, in the ``Disaster'' oracle, which symbolizes fire destroying a house, the original glyph places the fire within the house. However, the method without \(\mathcal{L}_{\text{GSDS}}\) mistakenly transforms the fire into a fired house. In contrast, accurately represents the fire and house components, our method effectively conveying the intended meaning of the glyph elements.

\begin{figure}[t]
    \centering
    \vspace{-1.5em}
    \includegraphics[width=1.0\linewidth]{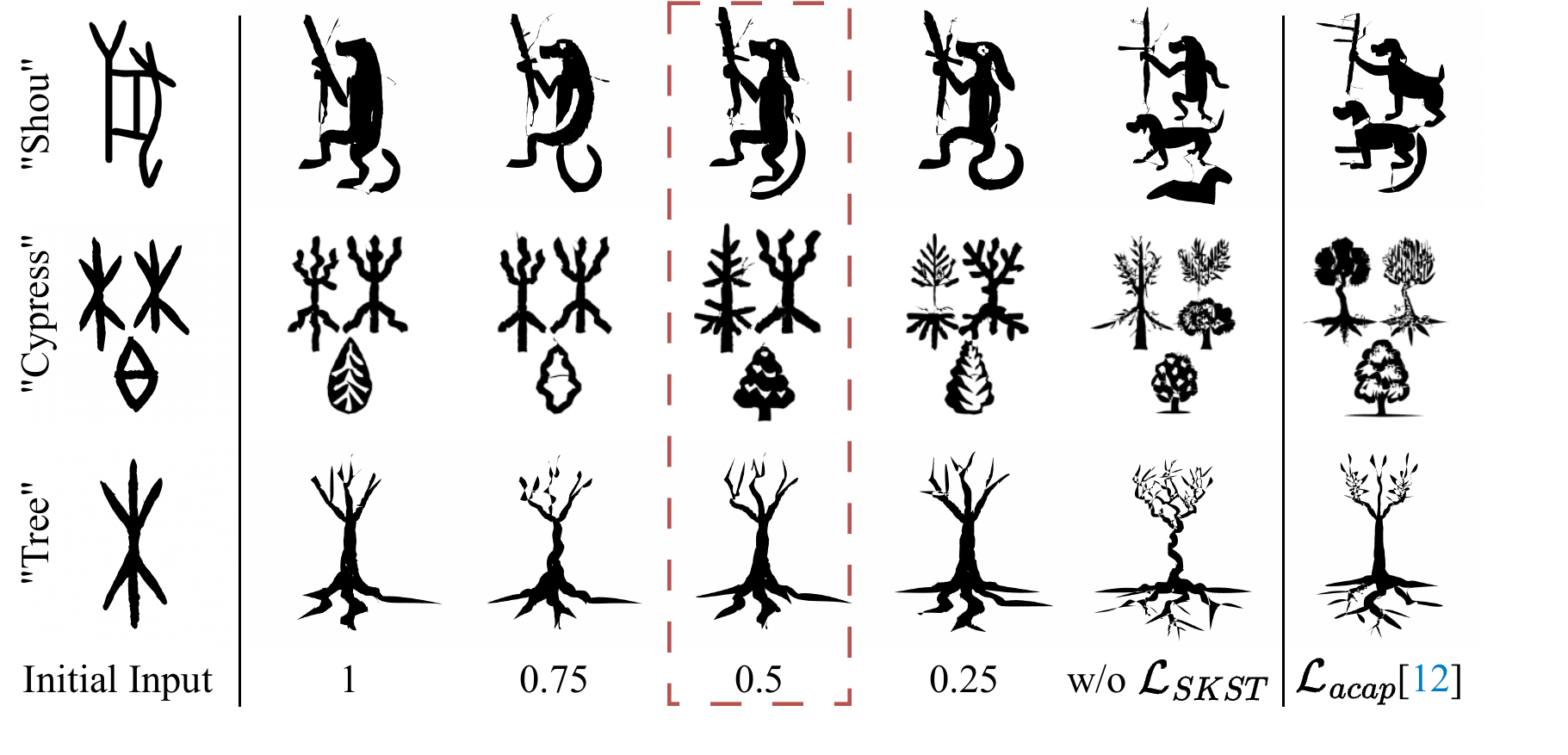}
    \vspace{-1.75em}
    \caption{Effect of the SKST loss and comparison of \(\mathcal{L}_{acap}\) loss. On the left to right are input glyph, the results from \(\mathcal{L}_{SKST}\) using \(\beta \in \{1, 0.75,0.5,0.25,0\}\), and results from \(\mathcal{L}_{acap}\)~\cite{wordasimage}. Note that images in red box best balances semantics and original glyph.}
    \vspace{-1.2em}
    \label{fig:ablation_sk}
\end{figure}

\noindent\textbf{Skeleton 
Strctrue Constrain. }Figure~\ref{fig:ablation_sk} illustrates the effect of weight \(\beta\) in the SKST loss on generated results, alongside a comparison with the ACAP loss from~\cite{wordasimage}. The results show that when \(\beta = 0\), the outputs from SKST resemble those from ACAP, indicating that ACAP struggles to preserve the initial shape with complex oracle structures. Additionally, when \(\beta\) is dominant, the SKST results closely resemble the original input, at the cost of losing semantic expression. Thus, to balance shape preservation and semantic expression, we select \(\beta = 0.5\) as the optimal parameter.

\subsection{Applications}
\begin{figure}[h]
    \centering
    \vspace{-1em}
    \includegraphics[width=1.0\linewidth]{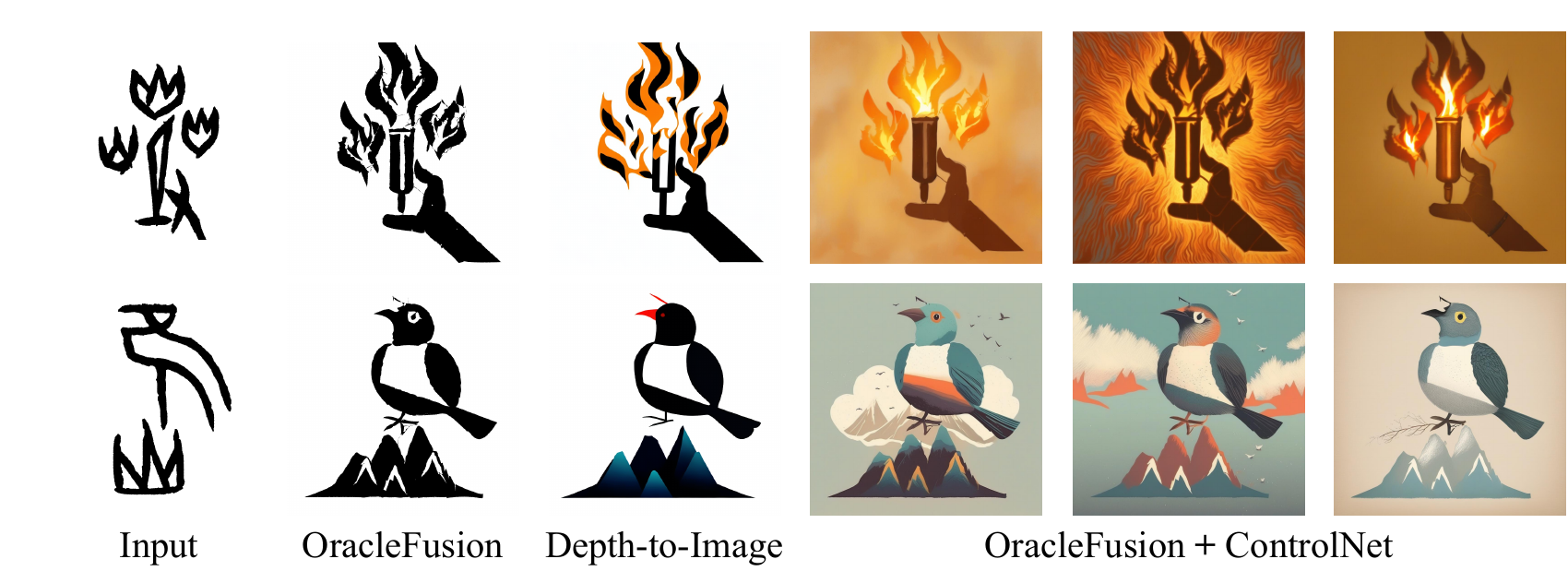}
    \vspace{-2em}
    \caption{Examples of utilizing D2I in SD2~\cite{sd} and Scribbles-to-Image with ControlNet~\cite{controlnet} as a post-processing step.}
    \label{fig:applications}
    \vspace{-2em}
\end{figure}

As shown in Figure~\ref{fig:applications}, OracleFusion integrates with D2I and S2I to enhance oracle bone script decipherment. The black-and-white glyphs generated by our model, enriched with semantic information, serve as structured inputs for these frameworks, enabling post-processing with color and texture. This integration improves visual fidelity and interpretability while highlighting OracleFusion’s role in bridging symbolic and computational decipherment.


%% file: sec/5_conclusion.tex
\section{Conclusion}

In this paper, we propose OracleFusion, a two-stage framework for generating semantically rich vectorized fonts to support in the interpretation of OBS. We utilize the MLLM to analyze the OBS data and propose a glyph structure-constrained semantic typography framework. Extensive experiments validate the framework’s generation and reasoning capabilities, demonstrating its effectiveness in offering novel insights for OBS interpretation and supporting decipherment. Future work will focus on expanding the multimodal dataset, enhancing OracleFusion's capabilities, and exploring broader applications in ancient script analysis.

%% file: supplement.tex
\clearpage
\setcounter{page}{1}
\maketitlesupplementary
\renewcommand{\thesection}
{\Alph{section}}
\appendix

\section{Applications of OracleFusion}
In this section, we explore additional applications of OracleFusion, emphasizing its potential to advance the decipherment of oracle bone script. The black-and-white glyphs produced by our model, enriched with semantic information, can be effectively employed as inputs for state-of-the-art frameworks such as the Depth-to-Image in Stable Diffusion 2~\cite{sd} and the Scribble-to-Image generation in ControlNet-SDXL 1.0~\cite{controlnet}. This integration highlights OracleFusion’s ability to generate structured and interpretable outputs, enabling downstream models to leverage these representations for image-based analysis and reconstruction tasks. These results further illustrate the versatility of OracleFusion as a foundational tool for bridging symbolic representation and computational decipherment methodologies.
\subsection{Depth-to-Image Post-Processing}
\begin{figure*}[t]
    \centering
    \includegraphics[width=1\linewidth]{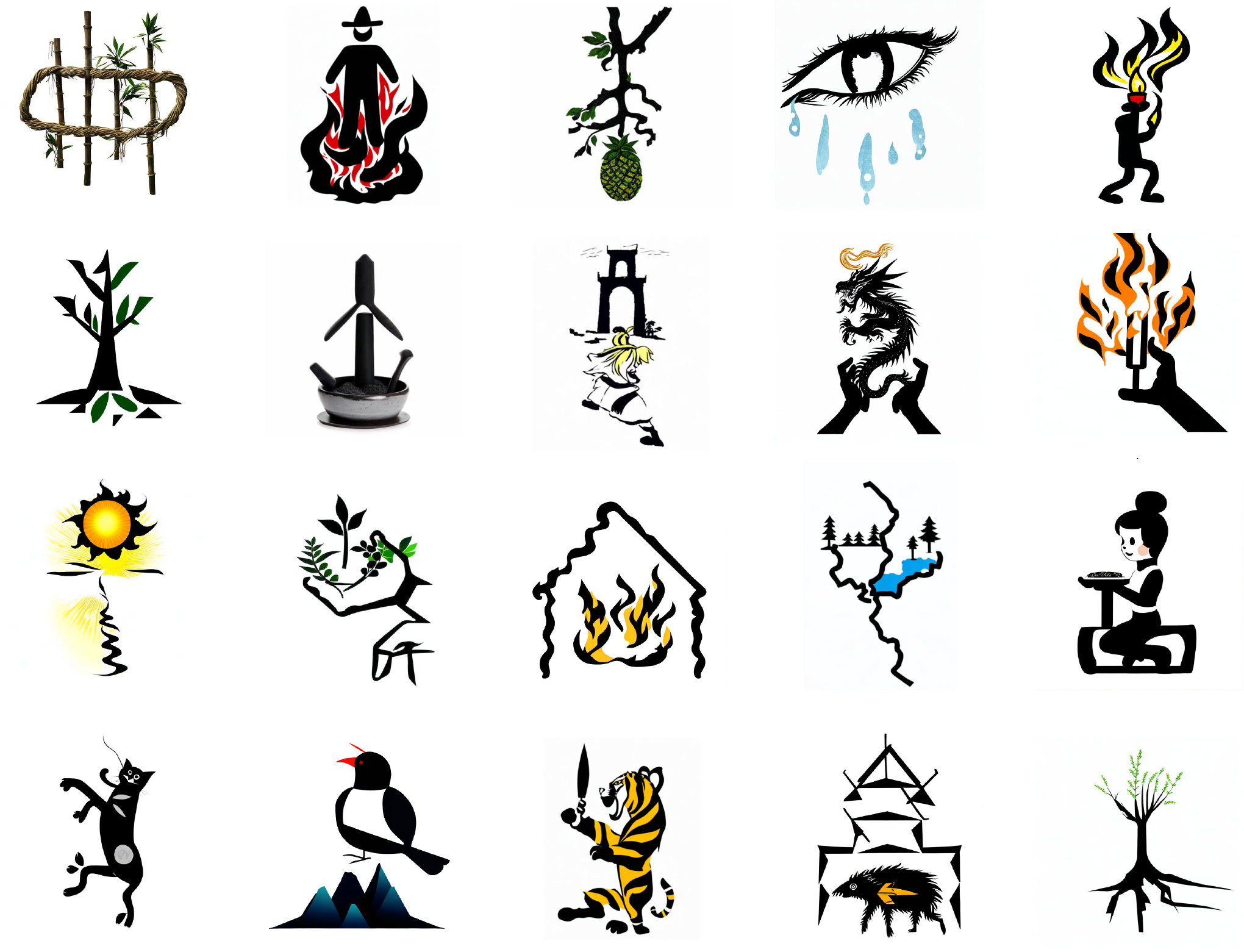}
    
    \caption{Examples of utilizing Depth-to-Image in Stable Diffusion 2~\cite{sd} as a post-processing step.}
    \label{fig:depth2image}
\end{figure*}

In this subsection, we employ the Depth-to-Image in Stable Diffusion 2~\cite{sd} as a post-processing step to enhance the outputs of our method by incorporating color and texture. The results, presented in Figure~\ref{fig:depth2image}, highlight the effectiveness of integrating depth-to-image processing for improving the visual fidelity and interpretability of the generated illustrations. This approach demonstrates the potential for combining generative and post-processing techniques to further refine and contextualize the outputs of computational decipherment systems.

\subsection{Scribbles-to-Image Post-Processing}
\begin{figure*}[t]
    \centering
    \includegraphics[width=1\linewidth]{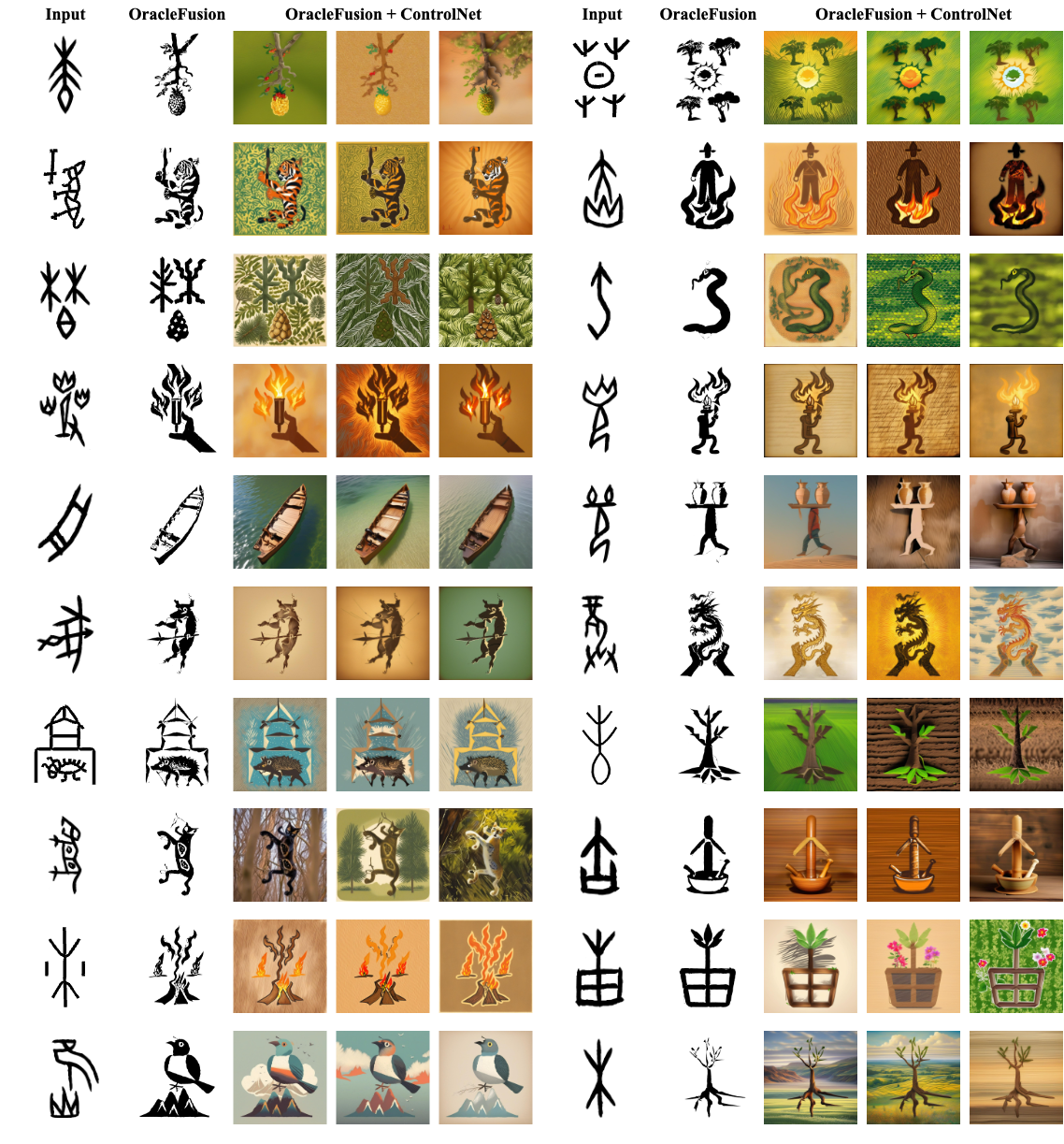}
    \caption{Examples of utilizing Scribbles-to-Image with ControlNet-SDXL 1.0~\cite{controlnet} as a post-processing step.}
    \label{fig:controlnet}
\end{figure*}

In this subsection, we utilize the results from our method as conditional inputs to guide the Scribbles-to-Image generation in ControlNet-SDXL 1.0~\cite{controlnet} to produce natural images. The results, presented in Figure~\ref{fig:controlnet}, demonstrate the potential of integrating this framework to significantly enhance the visual quality and realism of the generated illustrations. This integration highlights the power of combining structured semantic information with advanced generative models to improve the fidelity and interpretability of computationally generated content.
\section{Additional Decipherment Results}
\begin{figure*}
    \centering
    \includegraphics[width=1\linewidth]{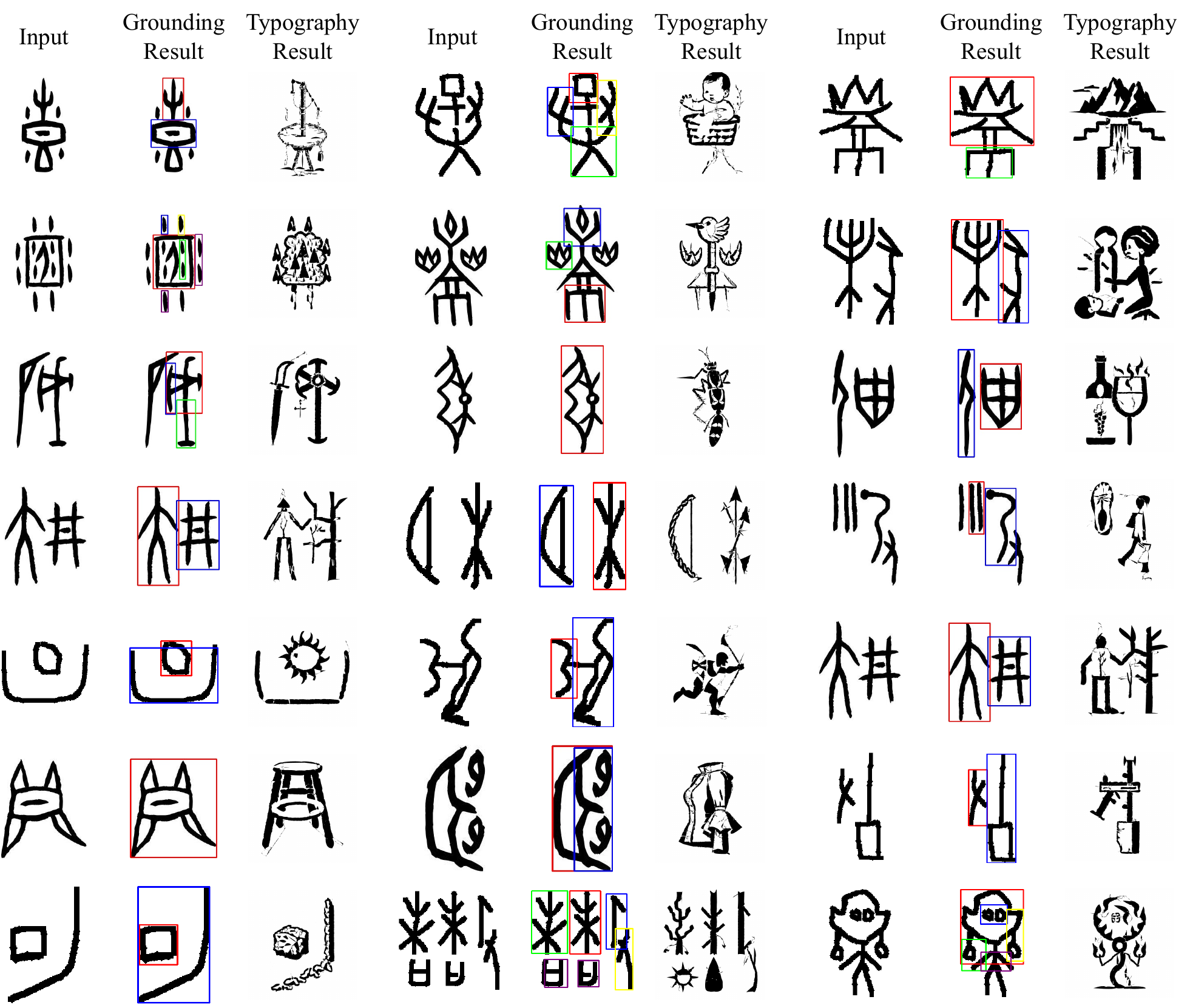}
    
    \caption{More grounding and semantic typography results for undeciphered oracle bone scripts. Note that the different colors of bounding boxes in the grounding results indicate distinct concepts of key structural components. The concepts are predicted by our method but omitted from the figure for clarity.}
    \label{fig:decipherment_result}
\end{figure*}

Figure~\ref{fig:decipherment_result} presents the decipherment results generated by OracleFusion for previously undeciphered oracle characters. For each character, the figure provides both grounding result and semantic typography result. Note that the different colors of bounding boxes in the grounding results indicate distinct concepts of key structural components. The concepts are predicted by our method but
omitted from the figure for clarity. This contribution aims to support scholars and researchers in advancing the computational study and linguistic analysis of ancient scripts.

\section{User Study Details}
\begin{figure*}[t]
    \centering
    \includegraphics[width=0.7\linewidth]{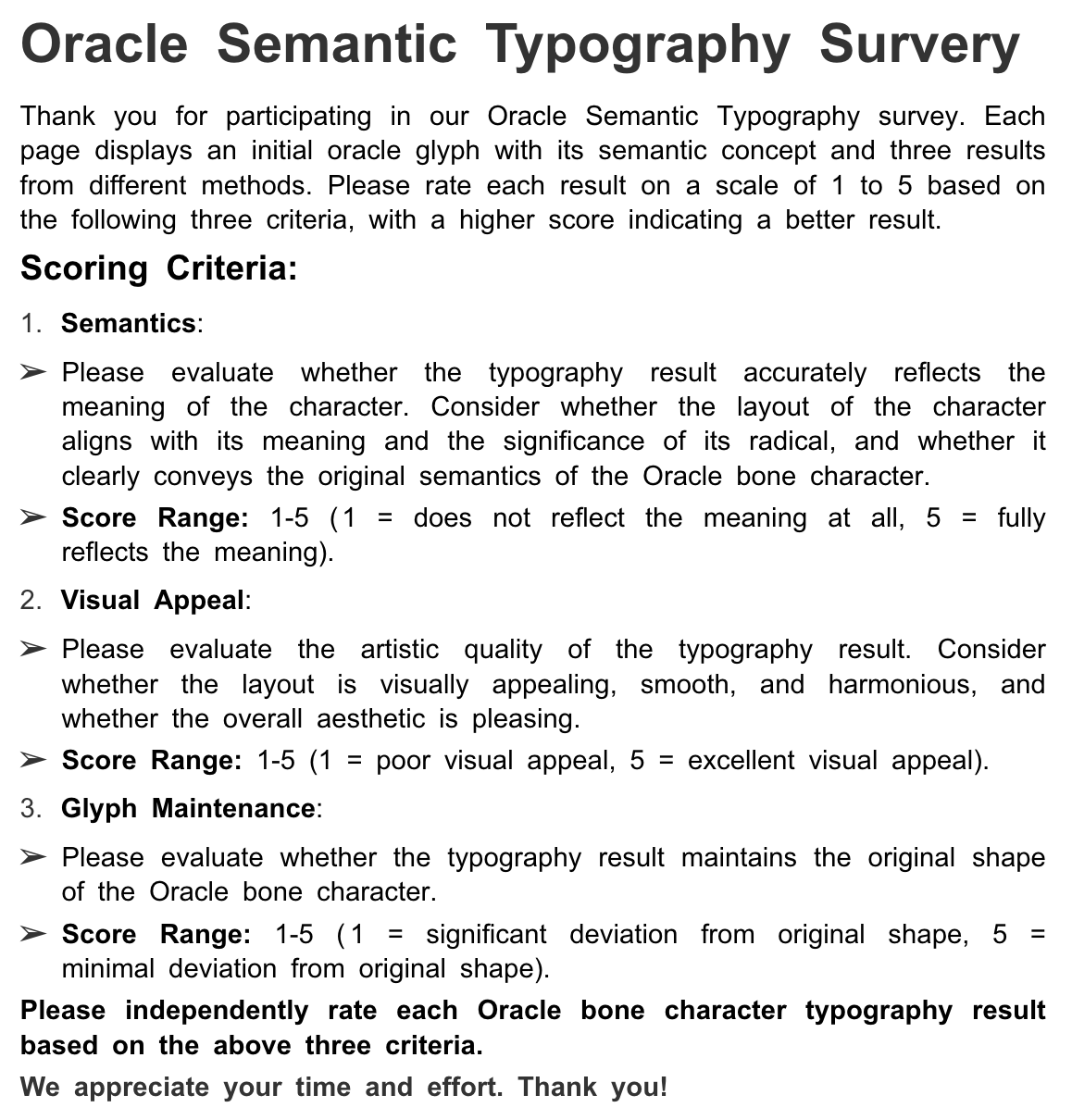}
    
    \caption{The description of our user study.}
    \label{fig:user_study}
\end{figure*}

\begin{figure*}[t]
    \centering
    \includegraphics[width=0.7\linewidth]{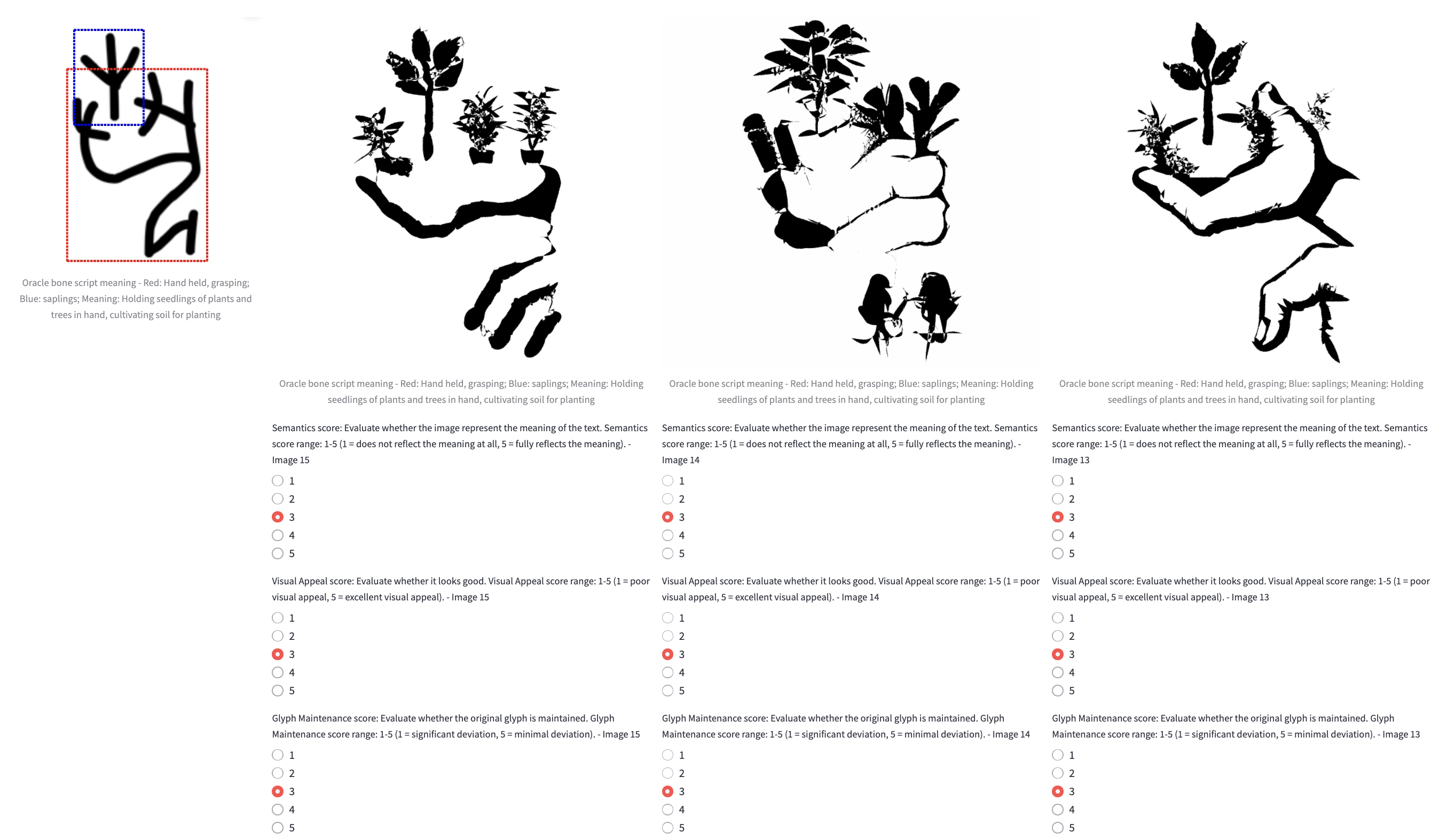}
    
    \caption{The description of our user study.}
    \label{fig:user_study_jiemian}
\end{figure*}

\begin{figure*}
    \centering
    \includegraphics[width=1\linewidth]{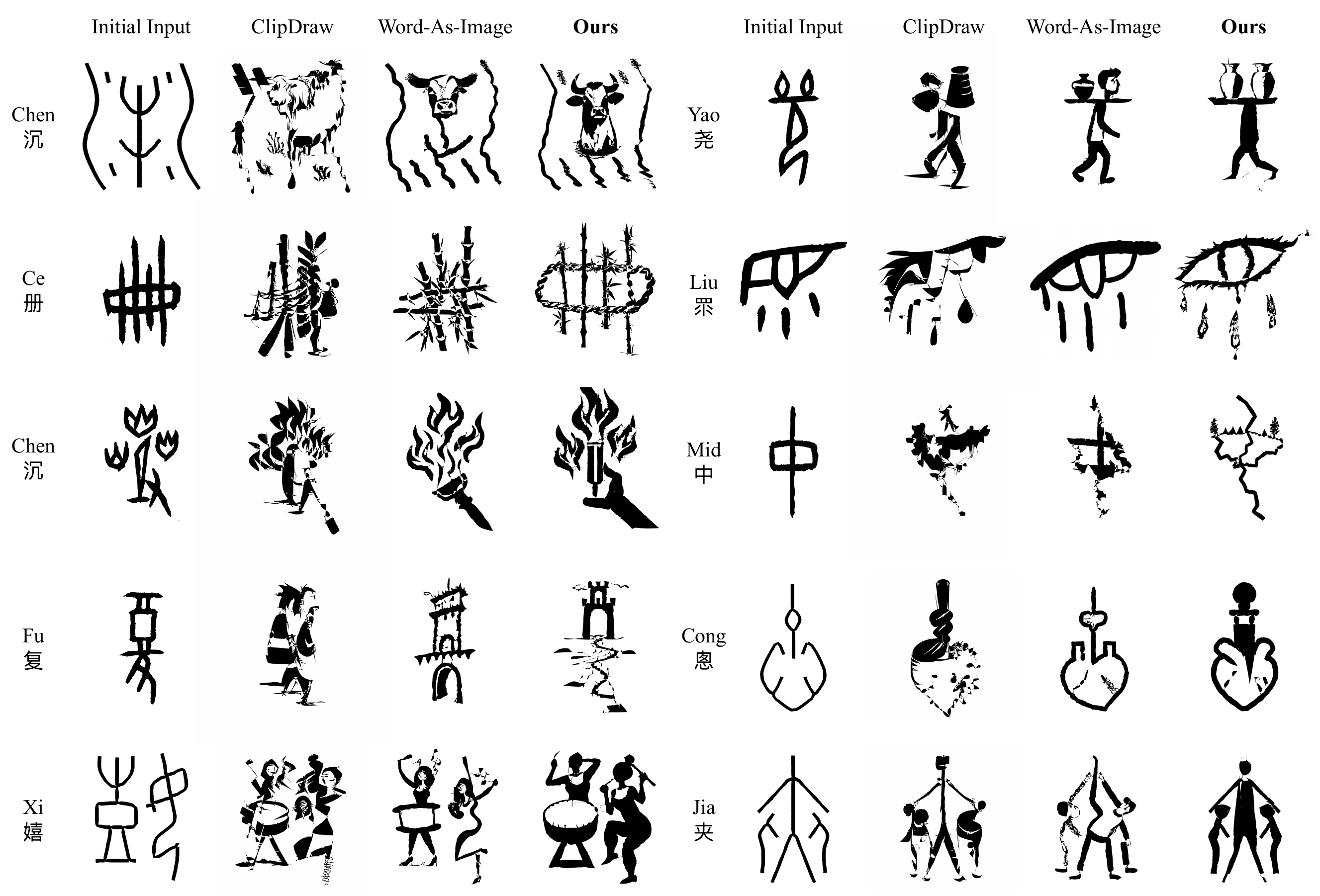}
    
    \caption{More qualitative comparison results between ClipDraw, Word-As-Image, and our OracleFusion in our user study.}
    \label{fig:user_study_compare}
\end{figure*}

To evaluate the effectiveness of our OracleFusion, we conducted a user study on comparisons between Word-As-Image~\cite{wordasimage}, ClipDraw~\cite{clipdraw} and our method. Figure~\ref{fig:user_study}
shows the description that participants read before answering the questions. As shown in Figure~\ref{fig:user_study_jiemian}, for each oracle character in our study, we present the initial glyph and three illustrations generated by the aforementioned three methods. Participants were asked to rate scores on semantics, visual appeal, and glyph maintenance, respectively. In Figure~\ref{fig:user_study_compare}, we display some of the qualitative comparison results in our user study. 



\clearpage
